# Generative Expressive Conversational Speech Synthesis

Rui Liu
imucslr@imu.edu.cn
Inner Mongolia University
Hohhot, China

Yifan Hu
22309013@mail.imu.edu.cn
Inner Mongolia University
Hohhot, China

Yi Ren
ren.yi@bytedance.com
ByteDance
Singapore, Singapore

Xiang Yin
yinxiang.stephen@bytedance.com
ByteDance
Shanghai, China

Haizhou Li
haizhouli@cuhk.edu.cn
SRIBD, School of Data Science
The Chinese University of Hong Kong
Shenzhen, China
National University of Singapore
Singapore, Singapore

## Abstract

Conversational Speech Synthesis (CSS) aims to express a target utterance with the proper speaking style in a user-agent conversation setting. Existing CSS methods employ effective multi-modal context modeling techniques to achieve empathy understanding and expression. However, they often need to design complex network architectures and meticulously optimize the modules within them. In addition, due to the limitations of small-scale datasets containing scripted recording styles, they often fail to simulate real natural conversational styles. To address the above issues, we propose a novel generative expressive CSS system, termed **GPT-Talker**. We transform the multimodal information of the multi-turn dialogue history into discrete token sequences and seamlessly integrate them to form a comprehensive user-agent dialogue context. Leveraging the power of GPT, we predict the token sequence, that includes both semantic and style knowledge, of response for the agent. After that, the expressive conversational speech is synthesized by the conversation-enriched VITS to deliver feedback to the user. Furthermore, we propose a large-scale Natural CSS Dataset called **NCSSD**, that includes both naturally recorded conversational speech in improvised styles and dialogues extracted from TV shows. It encompasses both Chinese and English languages, with a total duration of 236 hours. We conducted comprehensive experiments on the reliability of the NCSSD and the effectiveness of our GPT-Talker. Both subjective and objective evaluations demonstrate that our model outperforms other state-of-the-art CSS systems significantly in terms of naturalness and expressiveness. *The Code, Dataset, and Pre-trained Model are available at: https://github.com/AI-S2-Lab/GPT-Talker.*

## CCS Concepts

• **Information systems** → **Multimedia content creation**.

## Keywords

Conversational Speech Synthesis (CSS), User-agent Conversation, GPT, Expressiveness

## 1 Introduction

Conversational Speech Synthesis (CSS) (or Conversational Text-to-Speech (TTS), CTTS) aims to generate speech with proper style in the user-agent conversation scenario. In such scenarios, the user usually initiates a dialogue, then the agent and the user take turns to speak. During the interaction, the agent is expected to understand the user's needs and provide assistance and emotional support. Currently, with the increasing popularity of smart devices, there is a growing demand for human-machine interaction in various application scenarios such as smartphone assistants [47], smart home control [21], intelligent vehicle systems [15], and virtual reality / augmented reality interactions [13].

Many attempts on CSS have been proposed to enhance the naturalness and expressiveness of synthesized conversational speech from the perspective of context modeling. Guo et al [17] proposed a GRU-based context encoder to extract global prosodic information for the agent from the dialogue context. FCTalker [19] further incorporates the word- and sentence-level context knowledge, that represents the fine- and coarse-grained context, to enhance the context understanding ability of the agent. However, these works only consider the textual information, ignoring the audio and multi-modal dependencies in the conversation. To this end, researchers contributed to the study of multi-modal context modeling [10, 35, 39, 40]. However, they are increasingly inclined towards designing complex network architectures and optimizing the modules within them meticulously. For example, multi-scale multi-modal CTTS ($M^2$-CTTS) system [53] includes a textual context module and an acoustic context module with both coarse-grained and fine-grained modeling. Li et al. [28, 29] and Liu et al. [34] proposed the graph neural network based context learning schemes.

With the advancement of the Large Language Model (LLM), several studies aim to construct spoken language models that extend language models for the speech domain [23]. We have noticed that the Generative Pre-training Transformer (GPT) possesses concise and powerful context modeling capabilities [14] and has demonstrated impressive performance in capturing fine- and coarse-grained dependencies in tasks such as dialogue generation [58] and understanding [32, 33, 56]. Despite the successes, all the above works are not applicable to CSS, where multi-modal context is used as input to generate response speech output. We note that the context modeling of GPT aligns well with the requirements of CSS, yet it has been largely overlooked. Therefore, how to leverage GPT to build concise yet powerful context understanding solutions for user-agent interaction, will be the focus of this work.

In addition, the existing spoken dialogue dataset is challenging to meet the requirements for training GPT-based generative expressive CSS models in terms of both scale and quality. Specifically, $M^2$-CTTS [53], EmoSit-TTS [35] and ECSS [34], etc. all utilized a small-scale DailyTalk dataset [27], which contains 2541 dialogues audio in about 20 hours in total. More importantly, it only consists of recordings in a reading style, lacking sufficient expressiveness. Some works [10, 29] have attempted to use datasets such as IEMO-CAP [4] and ECC [28]. However, these datasets are designed for purposes like conversational emotion recognition [4, 41] and conversation education, and they may contain background noise issues that can affect the synthesis results. Therefore, it is necessary to construct a large-scale and high-quality expressive CSS dataset to support GPT-based CSS models in achieving a real natural conversational style.

In this paper, we propose a novel generative expressive CSS system, termed **GPT-Talker**. We transform the multimodal information of the dialogue history into discrete token sequences and seamlessly integrate them to form a comprehensive user-agent dialogue context. Leveraging the power of GPT, we predict the token sequence, that includes both semantic and style knowledge, of response for the agent. After that, the expressive conversational speech is synthesized by the conversation-enriched VITS to deliver feedback to the user. Furthermore, we propose a large-scale Natural CSS Dataset called **NCSSD**, that includes both naturally recorded conversational speech in improvised styles and dialogues extracted from TV shows. It encompasses both Chinese and English languages, with a total duration of 236 hours. The NCSSD dataset and related annotation will be public freely. We conducted comprehensive experiments on the reliability of the NCSSD and the effectiveness of our GPT-Talker. Both subjective and objective evaluations demonstrate that our model outperforms other state-of-the-art CSS systems significantly in terms of naturalness and expressiveness. In summary, our main contributions are as follows: 1) To the best of our knowledge, we are one of the earliest to introduce GPT into conversational speech synthesis and build a concise and powerful context modeling scheme for the user-agent conversation. 2) We have proposed a new large-scale Natural CSS dataset, termed NCSSD, that can support our GPT-Talker, even future GPT-style CSS model to achieve a real natural conversational speech style. 3) We have conducted comprehensive validations of the model's effectiveness and the dataset's reliability. Our model significantly outperforms baseline models in terms of naturalness and expressiveness.

## 2 Related Work
### 2.1 Conversational Language Model

Text and audio are two important modalities for human communications. Text-based LLMs have demonstrated remarkable achievements across various domains, including conversational chatbots [2], code generation [6], creative writing [46], and machine translation [36].

Inspired by the aforementioned works, many studies have recently explored conversational language modeling to address a variety of tasks involving speech and text [48, 49], such as automatic speech recognition [12], spoken question answering [1], and speech-to-text translation [51], etc. For instance, SpeechGPT [57] exhibited cross-modal conversational capabilities by employing discrete unit representations to convert continuous speech signals, and integrating LLMs with unit vocoder. This enables the model to effectively process multimodal input and generate corresponding output. dGSLM [38] proposed a dual-tower spoken LLM on discrete speech units to model two-channel spoken dialogue, but the generated spoken sentences lack semantic meaning. USDM [24] proposed a generalized speech-text pretraining scheme that leverages the chain-of-reasoning capabilities of LLMs to generate coherent spoken responses based on conversational speech. In human conversation, while the dialogue primarily relies on the lexical aspect, the speaking styles convey rich information beyond text, and can even alter the semantics of the spoken sentences [5]. To this end, E-Chat [52] proposed a novel LLM-based spoken dialogue system, that leverages an emotion embedding extracted by a speech encoder, enabling it to respond according to different emotional contexts. ParalinGPT [31] takes the conversational context of text, speech embeddings, and paralinguistic attributes as input prompts for LLM to improve current and response sentiment prediction, as well as response text generation, in natural human-human speech dialogue. Furthermore, Spoken-LLM [32] proposed to fuse the LLM and a self-supervised speech emotion representation model to help the LLM to predict response speaking style and text, enabling the subsequent expressive TTS model to generate natural and diverse speech responses.

Our work performs significantly differently from the abovementioned studies. Specifically, our work focuses on the task of CSS, where we integrate multi-turn multi-modal dialogue context into a unified sequence and predict the semantic and stylistic representations of the speech to be synthesized along with the current utterance. We then utilize this representation to generate the final expressive conversational speech. However, the aforementioned works, such as dGSLM, SpeechGPT, and USDM, overlooked effective mechanisms for modeling multi-turn dialogue history, while E-chat, ParalinGPT and Spoken-LLM etc. only generate text responses based on dialogue history rather than speech representations. Although MQTTS [7] and Pheme [3] also claim to be conversational speech generation systems, they do not model the dialogue context and only focus on the naturalness of the synthesized speech. Our GPT-Talker is similar to the recent style transfer TTS model, GPT-SoVITS [16], but an obvious difference is that GPT-SoVITS does not pay attention to the dialogue context information.

**Table 1: Comparison among other conversational datasets and NCSSD. ∗ means that the handling of collection data requires manual involvement. (EN: English; CN: Chinese; RC: Recording; CL: Collection.)**

| Dataset | Language | Source | Scale | |
|---|---|---|---|---|
| | | | Duration (h) | Speaker |
| IEMOCAP [4] | EN | RC | 12 | 10 |
| MELD [41] | EN | CL* | 13.66 | 407 |
| M³ED [59] | CH | CL* | 14.14 | 626 |
| CPED [8] | CH | CL* | 78.34 | 392 |
| ECC [28] | EN | CL* | 24 | 26 |
| DailyTalk [27] | EN | RC | 20 | 2 |
| STUDIES [44] | JP | RC | 8.2 | 3 |
| ASLP-CH [17] | CH | RC | 3 | 2 |
| RyanSpeech [54] | EN | RC | 10 | 1 |
| CALLS [43] | JP | RC | 6.5 | 1 |
| **NCSSD (Ours)** | EN, CH | RC, CL | 236 | >776 |

## 2.2 Conversational Speech Datasets

Table 1 provides a summary of the existing relevant datasets to our knowledge. Lines 6-11 present information about some conversational speech datasets specifically designed for CSS task. It can be observed that these datasets are relatively small in scale. For example, ECC [28] gathered 66 sets of public videos from YouTube's English Conversation channel, amassing 24 hours of content. These dialogues involve two, three, or multiple participants. DailyTalk [27] is derived from DailyDialog dataset [30]. It showcases 2,541 high-quality English conversations between a male and female, spanning a total of 20 hours. The Japanese corpus for empathic conversations STUDIES [44] and CALLS [43] covers two scenarios, that are *communication between teachers and students in school* and *customer service via telephone*, featuring 8.2 and 6.5 hours of speech, respectively. RyanSpeech [54] is a high-quality male TTS corpus in the conversational domain, that contains 9.84 hours audio samples. Note that CALLS and RyanSpeech claim to be conversational datasets, but they only contain one speaker and are not suitable for CSS task. Guo et al. [17] recorded an internal dataset (we call "ASLP-CH" here) comprising 3 hours of conversations between 2 Chinese women, who assumed the roles of a customer and a customer service representative, respectively.

In summary, the scale of these datasets is insufficient to train a high-quality CSS model based on the GPT model. This necessitates the creation of a large-scale, freely available CSS dataset. Additionally, our dataset has advantages in terms of *language diversity* and *data source diversity* compared to existing datasets. Specifically, it includes bilingual data in both Chinese and English, as well as subsets of recorded data and collected data. Furthermore, we have developed an automated processing pipeline for collected data, greatly improving the efficiency of creating natural CSS datasets. We hope that this initiative can contribute to the development of the community. Furthermore, in Table 1, we have listed some of the latest dialogue speech datasets designed for tasks such as dialogue emotion understanding. It can be seen that our data still has significant advantages in terms of language diversity, data source diversity, and dataset scale compared to these datasets.

## 3 GPT-Talker: Methodlogy

### 3.1 Task Definition

In user-agent spoken conversation, the user and the agent take turns speaking. The user speaks first, and the agent understands the user's semantics to provide a spoken response. As time progresses, multi-turn user-agent dialogue history accumulates and forms. The task setting in our CSS task, therefore is to generate the speech of agent's response according to the conversation context, where the text of the response is given. Specifically, assume that the $N$-turn conversation context includes the multi-modal dialogue history $\mathcal{H}$ and the current utterance $C$, where $\mathcal{H} = \{(U_1^t, U_1^a, S_1), (U_2^t, U_2^a, S_2)..., (U_{N-1}^t, U_{N-1}^a, S_{N-1})\}$ and $C = (U_N^t, S_N)$. ($t$ and $a$ means the text and audio modalities respectively, $S$ is the speaker identify label.) The goal of CSS is to ensure that the synthesized speech $U_N^S$ is suitable for the whole dialogue situation. Therefore, how to model the multi-turn multi-modal conversational context and provide a proper speaking style for the agent is the focus of the task.

To this end, our GPT-Talker consists of two key components, that are *GPT-based context modeling* and *Expressive conversational speech synthesis*. Specifically, GPT-based context modeling proposes a novel Conversation GPT (ConGPT) to model the multi-turn multi-modal conversational context by treating the discrete token sequence of context as the condition prompt, and predict proper semantic and style expression for the agent. Expressive conversational speech synthesis proposes the Conversational VITS (ConVITS) to enrich the VITS with the agent's semantics, style, and timbre to generate expressive conversational speech based on the known response content.

### 3.2 Conversational GPT

As shown on the left side of Fig. 1, The ConGPT encompasses 1) Multi-turn Multi-modal Context Tokenization and 2) ConGPT-based Semantic and Style Inference. The former module converts the multi-modal context into a unified discrete sequence, and then constructs a discrete representation of the multi-turn conversation context. Based on this representation, The latter module infers the semantics and style of the response speech.

*3.2.1 Multi-turn Multi-modal Context Tokenization.* Unlike traditional CSS works that adopt complex graph neural networks [29, 34] or cascade pipelines [35] to understand the context, our ConGPT seeks to understand the multi-modal context within a unified discrete sequence directly.

**Textual Tokenization**: Similar with VALL-E [48], we convert the text data of dialogue history $\mathcal{H}$ and current utterance $C$, including $U_{1 \to N-1}^t$ and $U_N^t$, into the discrete phoneme sequences. As shown in Fig. 1, the discrete phoneme sequences of $U_{1 \to N}^t$ are represented by $T_{1 \to N}^t$. The phoneme encoder is built based on the g2p_en

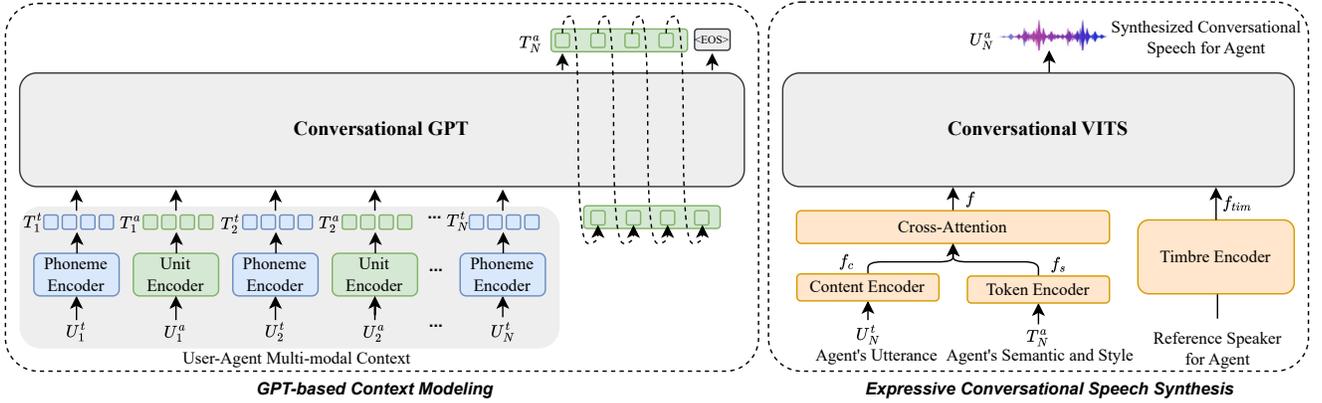

Figure 1: The overview of GPT-Talker, that includes the Conversational GPT and the Conversational VITS.

python module [1] for English and the opencpop-strict dictionary [2] for Chinese.

**Acoustic Tokenization**: To model natural speech conversations, the discrete speech representation must contain not only the semantics of the utterance but also expressiveness features such as speaking style, which are crucial for CSS task. Previous research [24] has demonstrated that HuBERT tokens contain rich semantic information as well as strong traces of paralinguistic features that can be used to accurate classify speech emotions. Therefore, we follow [24] and employ HuBERT [18] to acquire high-dimensional speech representations. Inspired by [26], we do not employ unit deduplication on HuBERT units, since it maintains a consistent length ratio with the speech, removing the need to prepare a separate duration predictor. Specifically, we adopt speech units as acoustic context units that are derived from k-means clustering of the HuBERT's intermediate speech representation. We follow GPT-SoVITS and then adopt a Vector Quantization (VQ) layer to convert the acoustic context units to the learnable token sequence, thus better matching the expressive speech generation task. In this way, the audio data of dialogue context $U^a_{1 \to N-1}$ are represented by $T^a_{1 \to N-1}$.

**Unified Context Serialization**: After acquiring discrete tokens for text and audio modalities of context, we emulate the real-time dialogue flow and combine the textual and acoustic tokens alternately into a multi-modal conversational format as $\{T^t_1, T^a_1, T^t_2, T^a_2, ..., T^t_N\}$. We adopt an alternating approach between the text and speech modalities instead of serializing the conversation history by first combining the text modality and then concatenating the speech modality. We believe that the alternating approach can better capture the cross-modal contextual dependencies in the conversation history.

*3.2.2 ConGPT-based Semantic and Style Inference.* To inference the Semantic and Style information of response speech according to the serialized multi-modal user-agent context. the ConGPT takes the entire serialized conversation context $\{T^t_1, T^a_1..., T^t_{N-1}, T^a_{N-1}, T^t_N\}$ as input, then predicts the semantic and style knowledge $T^a_N$. The process is formulated as follows:

$$T^a_N = ConGPT(\{T^t_1, T^a_1, T^t_2, T^a_2..., T^t_N\}) \quad (1)$$

[1]https://pypi.org/project/g2p-en/
[2]https://wenet.org.cn/opencpop/

where $T^a_N$ also follows the same tokenization method as $U^a_{1 \to N-1}$ to discretize $U^a_N$ in advance, serving as the output of ConGPT. The sub-tokens in the discrete token sequence $T^a_N$ are decoded using an autoregressive approach until the End of Sequence (EOS) label is decoded. Please note that, similar to [33], we do not assign explicit speaker identities to the token sequence of the serialized context.

### 3.3 Conversational VITS

After unified context modeling, the Conversational VITS (ConVITS) is proposed to provide proper expressive speech for the agent's response. Note that ConVITS takes three information sources of the agent, including the agent's utterance $U^t_N$ for content rendering, the semantic and style representations of the agent inferred by the ConGPT for semantic and style rendering, and the additional reference speaker for timbre rendering. As shown in the right panel of Fig. 1, it consists of four key complements, which are the context encoder, token encoder, timbre encoder and the VITS synthesizer. The cross-attention mechanism is used to integrate the content, semantic and style information of agent.

*3.3.1 Content, Semantic and Style Rendering.* First, we utilize the text encoder $Encoder_t(\cdot)$, that consists of 6 layers of transformer encoder, to process the response content $U^t_N$, extracting its inherent textual information $f_c$ to ensure the intelligible of synthesized speech. After that, the predicted context-aware semantic and style tokens $T^a_N$ are converted into a high-level style representation $f_s$ using the token encoder $Encoder_s(\cdot)$, which consists of 3 layers of transformer encoder.

Finally, to achieve unified content, semantic and style rendering for expressive CSS, textual information $f_c$ and high-level style representation $f_s$ are integrated into a final agent's style embedding $f$ via the cross-attention layer. We use a multi-head attention module in the cross-attention layer, which accommodates different input lengths. Here, we treat $f_s$ as the query and $f_c$ as both the key and value.

*3.3.2 Timbre Rendering.* In order to achieve flexible and diverse expressive styles, we do not rely on fixed speaker IDs for speaker control. Instead, we employ a timbre encoder with a reference encoder to perform timbre transfer on any reference speech, allowing for flexible speaker timbre rendering based on content, semantics, and style rendering. As shown in the Fig. 1, the timbre encoder $Encoder_{tim}(\cdot)$, consists of 6 convolutional layers, a GRU layer, and a linear layer, takes a reference speech $U_{agent}^a$ for a specific speaker and extracts the timbre embedding $f_{tim}$.

During training, the speech $U_{N-2}^a$ from the agent in the last turn is chosen as the reference speech $\tilde{U}_{agent}^a$. During the inference stage, we can assign a reference with any speakers to achieve zero-shot timbre rendering performance.

*3.3.3 Speech Generation.* The speech generation stage extends the vanilla VITS [25] into conversational settings by leveraging the above conversational expressiveness rendering. It's important to highlight that the ConVITS architecture excludes the Stochastic Duration Predictor, since the semantic and style tokens predicted by ConGPT already encompass duration information, there's no necessity to predict it separately.

The final agent's style embedding $f$ is processed by a projection layer to derive the mean $\mu$ and variance $\theta$. The agent's timbre embedding $f_{tim}$ is encoded by the posterior encoder to produce the latent normal posterior distribution variable $z$, which is then decoded by a Flow-based Decoder to generate the normalized stream $f_\theta(z)$. The HiFi-GAN generator then upsamples the latent variables $z$ to the speech waveform $U_N^a$.

## 3.4 Training Strategy

We propose a **Three-Stage** training strategy to ensure the performance of our GPT-Talker: 1) In the first stage, we focused on the modeling capabilities of ConGPT and ConVITS in single-sentence speech scenarios. we trained ConGPT and ConVITS with single-sentence speech datasets including LibriTTS [55], LJSpeech [20], AISHELL-3 [45], etc., that the total duration is about 2.5k hours. In this way, ConGPT can predict the speech tokens based on the text token sequence. The ConVITS performs stable speech generation after inputting the provided text. 2) In the second stage, we continue to train the ConGPT using the collection subset of NCSSD. This enabled it to accurately predict the semantics and stylistic knowledge of the current sentence, leveraging the provided dialogue context and multimodal information. 3) In the third stage, we further enhance the naturalness and expressiveness of the synthesized speech by fine-tuning both ConGPT and ConVITS using the recording subset of NCSSD.

Concerning the computation of the model's total loss. The GPT-Talker's loss $L_{total}$ is composed of two elements: $L_{ConGPT}$ and $L_{ConVITS}$. $L_{ConGPT}$ calculates the cross-entropy loss between the predicted acoustic units and the real units. $L_{ConVITS}$ includes mel reconstruction loss $L_{recon}$, KL divergence loss $L_{vae}$ from VAE, feature-matching loss $L_{fm}$, adversarial training loss $L_{adv}$, and $L_{vq}$ from Vector Quantization. Specifically, $L_{vq}$ computes the commitment loss between the quantized vector and the input discrete token.

## 4 NCSSD Construction

As mentioned before, we also propose a large-scale natural conversational speech corpus, **NCSSD**, to support the GPT-based CSS training. As illustrated in Table 1, the dataset includes two subsets, that are the collection part and the recording part, covering a total duration of over **236 hours**. Our dataset is available with the CC-BY-SA 4.0 license. Unlike the traditional data production method with human participation [8, 41, 59], note that we propose an automatic data construction pipeline for the collection subset and a ChatGPT-assisted workflow for the recording subset. We will report the data collection process [3] and the data statistics results.

### 4.1 Automatic Data Construction for Collection Subset

The automatic data construction pipeline for the collection subset consists of 1) Video Selection, 2) Dialogue Scene Extraction, 3) Dialogue Segment Extraction, and 4) Dialogue Script Recognition.

*4.1.1 Video Selection.* In order to build a large-scale, diversified, and natural conversational dataset, we collect videos from different TV series, which can simulate spontaneous conversation behavior in the real-world environment [41, 59]. We collect 79, and 34 TV shows for English and Chinese respectively. The detailed list of all TV series are introduced in Appendix. Unlike other data creation methods that require manual selection of TV shows based on certain rules [59], we automatically filter out eligible dialogues and corresponding timestamps using *Dialogue Scene Extraction*, *Dialogue Segment Extraction*, and *Dialogue Script Recognition* modules.

*4.1.2 Dialogue Scene Extraction.* A complete TV show consists of multiple dialogue scenes (may include two or more speakers) that are interconnected but independent from each other. To extract these dialogue scenes, we employ Voice Activity Detection (VAD) technology, which uses silent segments in the audio information of the entire TV show to identify dialogue scenes. Subsequently, the extracted dialogue scene audio is further processed to obtain clean dialogue speech.

Specifically, we first employ a pre-trained VAD model, silero-vad[4], to identify the timestamps of non-silent voice chunks in the video, since the silero-vad was trained on huge corpora that include over 100 languages and it performs well on audios from different domains with various background noise and quality levels. We set the silent segments threshold to 4 seconds to get the VAD results, since 4 seconds of silence often indicates the start of a new dialogue, as proven by extensive preliminary data testing. Then we segment the complete audio file of a video into various discontinuous audio clips that represent the various dialogue scenes. To ensure that each dialogue scene includes multi-turn dialogues with appropriate length, we further discard the audio clips where "the ratio of silent to non-silent segments exceeds 30%" or the "total duration is less than 15 seconds".

Subsequently, we then perform background music separation with Demucs [5] to discard the background music and other distracting information. To further refine the vocal component's quality, we thoroughly assess the signal-to-noise ratios (SNR) of both vocal and background noise, retaining only vocal chunks with an SNR above 4, and adopt speech enhancement model, sepformer [6], to obtain the clean audio for all dialogue scenes.

*4.1.3 Dialogue Segment Extraction.* Dialogue segment extraction aims to extract the conversational speech containing only two-person interaction from the previously obtained dialogue scene speech.

---
[3]For a more intuitive flowchart, please refer to the Appendix.
[4]https://github.com/snakers4/silero-vad
[5]https://github.com/facebookresearch/demucs
[6]https://huggingface.co/speechbrain/sepformer-dns4-16k-enhancement

To achieve this, we utilized a speaker recognition interface [7] based on ByteDance to analyze the speech information from all dialogue scenes. This interface provides us with numerical speaker labels and corresponding timestamp information. Subsequently, we extract dialogue segments for two-person conversations based on the speaker labels and timestamps. Our extraction criteria required the dialogue segment to have more than four utterances, with each speaker contributing at least two utterances.

*4.1.4 Dialogue Script Recognition.* Through the above steps, we obtained the dialogue-level audio and visual information of the two-person conversation. In order to obtain the dialogue script, we employ the Alibaba automatic speech recognition engine [8] to recognize the speech. Finally, the audio, scripts and the visual information of all dialogues are combined tighter as the collection suset of NCSSD.

## 4.2 ChatGPT-assisted Data Construction for Recording Subset

Unlike the collection subset, the construction of the recording subset involves designing dialogue scripts and inviting volunteers to record their voices and upper body image signals. The script serves as a prompt for the dialogue content rather than a strict guideline, and volunteers are allowed to spontaneously expand on the dialogue during the recording process. The resulting dialogue speech, the upper body image and the script are then recorded to obtain the final recorded data.

*4.2.1 Dialogue Script Draft Generation.* The first step in preparing recorded speech data is to create dialogue scripts. Manual preparation of scripts can be time-consuming, labor-intensive, and may lack diversity in terms of dialogue topics. Therefore, we leverage the powerful text generation capabilities of ChatGPT by using a prompt-based approach to generate diverse dialogue scripts that meet our expectations.

Specifically, we employ GPT-3.5 Turbo version of ChatGPT to generate the script of a two-person conversation. As shown in Fig. ??, we design a two-step prompt template to prompt the ChatGPT to output large-scale dialogue scripts to reflect spontaneous conversation behavior in the real world. In the first step, we aim to prompt ChatGPT to generate various topic words of human conversation ensuring the richness of communication content. In the second step, we select a specific dialogue topic and set the emotional state of the starting speaker. We prompt ChatGPT to generate multi-turn dialogues with a range of 4 to 15 turns. Additionally, we request GPT to add emotion and intent labels [9] to the generated scripts to enhance their interpretability and provide more reference information for subsequent speech recordings. The detailed prompt templates are shown in Appendix.

*4.2.2 Spoken Dialogue Recording.* Based on the previously obtained scripts, we invited volunteers to record the dialogue speech. We also captured the upper body images of the participants during the recording. Importantly, during the recording process, volunteers are encouraged to freely add dialogue content based on the emotional and intent information provided in the script, in order to create spontaneous and natural dialogue speech.

Table 2: The detailed data statistics of NCSSD.

| Item | Collection | | Recording | |
|---|---|---|---|---|
| | EN | ZH | EN | ZH |
| Language | *EN* | *ZH* | *EN* | *ZH* |
| Dialogues | 7,033 | 8,776 | 1,196 | 2,451 |
| Utterances | 62,603 | 99,126 | 10,033 | 21,688 |
| Words | 856,011 | 1,688,778 | 157,967 | 507,008 |
| Min words per utterance | 1 | 1 | 2 | 2 |
| Max words per utterance | 299 | 322 | 53 | 92 |
| Duration(h) | 72.94 | 115.22 | 19.10 | 29.57 |
| Max dialogue duration(s) | 177.51 | 308.21 | 108.52 | 75.82 |
| Min dialogue duration(s) | 4.74 | 5.54 | 20.06 | 15.73 |
| Mean utterance duration(s) | 4.19 | 4.18 | 6.85 | 4.90 |
| Max dialogue turns | 39 | 69 | 15 | 14 |
| Mean dialogue turns | 8.90 | 11.29 | 8.38 | 8.84 |
| Min dialogue turns | 4 | 4 | 5 | 6 |
| Speakers | > 339 | > 410 | 11 | 16 |

Specifically, we employed 27 young volunteers with English as their second language [10] to participate in the recording sessions. Their compensation is based on the number of dialogues recorded. The recording venues varied, including classrooms, meeting rooms, seminar halls, and more, providing a diverse range of settings. We allow the volunteers to engage in spontaneous dialogues guided by the provided script, emotion label, and intent label. This approach ensured that the final recorded voice sounded natural and authentic.

*4.2.3 Dialogue Script Re-identification.* Due to the presence of spontaneous utterances by the voice actors during the recording, we use a speech recognition interface [8] to re-transcribe the recorded speech and obtain the recognized text as the final dialogue script. Together with the recorded speech and upper body images, this forms our final recording subset.

## 4.3 Data Statistics

Table 2 presents the overall statistics of the NCSSD dataset. It contains a total of 19,456 dialogues and 193,450 sentences from 113 TV shows, ensuring the scale and diversity of the data. The dataset spans approximately 236 hours, featuring the longest Chinese conversation at 308.21 seconds and the longest English conversation at 177.51 seconds. Every conversation exceeds 4 seconds, satisfying the minimum duration needed for dialogue tasks. On average, each conversation contains at least 8 turns, facilitating the training of models for extended dialogue sequences. With over 776 speakers, the dataset encompasses a diverse range of speaking styles and habits. For additional statistics of the NCSSD, please refer to Appendix.

---

[7] https://www.volcengine.com/product/voice-tech
[8] https://ai.aliyun.com/nls/
[9] The emotion categories are labeled using a 7-category scheme [59], that are Neutral, Happy, Surprise, Fear, Angry, Disgust, and Sad. The intention labels are labeled using a 9-category scheme [50], including Question, Agree, Acknowledge, Sympathize, Encourage, Console, Suggest, Wish, and Neutral.

[10] Due to budget limitations, we did not invite native English speakers for the recording. Although the language proficiency of our participants may not be their native language, they possess fluent English listening, speaking, reading, and writing skills. We believe they are competent enough to carry out our data recording tasks.

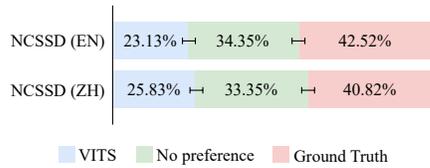

**Figure 2: The ABX preference test results between VITS and Ground Truth (GT) speech on our NCSSD.**

## 5 Experiments

In the experiment part, we will first assess the quality of NCSSD in a single-sentence speech synthesis scenario. Subsequently, we validate the GPT-Talker by comparing it with various state-of-the-art baselines of CSS task. We will introduce the baselines, and evaluation metrics in the following subsections and report the results in the next section. Note that the experimental setup can be viewed in the Appendix.

### 5.1 Baseline Models

To validate the reliability of NCSSD, we evaluate it on the single-sentence speech synthesis task and select the advanced end-to-end TTS model, VITS [25], as the baseline. Note that VITS was also chosen as the backbone network by many CSS models [9, 11], and we believe it is appropriate to serve as the baseline.

To evaluate the validity of GPT-Talker in terms of context understanding and modeling, we develop three advanced CSS systems that represent various context learning methods with the following three categories: (1) **GRU-based Context Learning**: The conversational context-aware TTS (we call "CCATTS" here) model proposed by [17] employs a GRU-based network to model the sentence-level dependency among the dialogue context; (2) **Multi-scale Context Learning**: FCTalker [19] is an representative work that consider both the sentence-level and word-level contextual within the dialogue context; (3) **Heterogeneous Graph-based Context Learning**: ECSS [34] is an advanced expressive and emotional CSS model that adopts heterogeneous graph to model the complex relation among the multi-modal context.

### 5.2 Metrics

We conduct comprehensive evaluations using both objective and subjective metrics.

**Subjective Metrics.** The subjective evaluation includes three components: (1) **ABX preference test**, where listeners had to decide which of the transformed speeches, A or B (produced by two different methods), sounded closer to ground truth speech or if they had no preference. (2) **Dialogue-level Mean Opinion Score in terms of Naturalness (N-DMOS)**, which requires participants to primarily assess its naturalness under the dialogue context. (3) **Dialogue-level Mean Opinion Score in terms of Emotion (E-DMOS)**, which primarily focuses on assessing the emotional expression conveyed through the speech and determining its alignment with the ongoing emotional tone of the dialogue context.

**Objective Metrics.** (1) Dynamic Time Warping Distance **DTWD**: we follow [42] and adopt average dynamic time warping (DTW) distance [37] of the pitch distribution for the ground-truth speech and synthesized speech to evaluate expressiveness-related performance. (2) **Speaker Similarity (SSIM)**, we follow [22] and utilize the speaker verification model, that fine-tuned with WavLM[11], to extract the speaker embedding for the ground-truth speech and the synthesized speech. After that, we calculate the cosine similarity score between two embeddings as the final speaker similarity metric. The similarity score ranges from [-1, 1], where a higher value indicates a greater similarity.

## 6 Results and Discussions

We report the results of the following aspects, including "Reliability verification of NCSSD", "Validity verification of GPT-Talker" and "Analysis of Three-Stage Training". Please refer to the Appendix to check more results.

### 6.1 Reliability Verification of NCSSD

We train VITS using our NCSSD in single-sentence scenarios, and the ABX preference test results of synthesizing Chinese and English speech are shown in Fig. 2.

For each language, we combine the collection and recording subsets to train the VITS model. We randomly select 50 sentences from both the Chinese and English datasets for synthesis and conduct an ABX preference test with 30 volunteers to mark the speech that has higher naturalness between the speech synthesized by the VITS model and the ground truth (GT) speech. The results of Fig. 2 indicate that the performance of VITS is very close to that of the GT. The proportion of "No Preference" in Chinese speech reaches as high as 33.35%, while in English speech, it is approximately 34.35%. This demonstrates that our NCSSD can support advanced TTS model in synthesizing high-quality speech.

Undoubtedly, the performance of GT speech is expected to be better. However, since our NCSSD is designed to mimic real-world human conversations, effectively incorporating contextual modeling methods is an effective approach to further enhance the performance of TTS models. Therefore, in the next section, we further validate the effectiveness of the proposed GPT-Talker in context learning.

### 6.2 Validity Verification of GPT-Talker

In this section, we will compare GPT-Talker with three advanced CSS baselines. Since these baselines have conducted experiments using the English dataset DailyTalk in their respective studies, we will validate our model using the English portion of the NCSSD dataset and the DailyTalk dataset [12]. Similar to the previous section, the English portion will be trained by combining the collection subset and the recording subset.

For subjective evaluation, due to budget constraints, we don't invite native English speakers as volunteers. Instead, we invite 30 Chinese students who are proficient in English listening, speaking, reading, and writing as volunteers. We randomly select 50 sentences from the test set to synthesize speech for each system to be tested. Then, we ask the volunteers to rate the speech using N-DMOS and E-DMOS based on the guidelines. Their compensation was calculated based on the number of test samples. Additionally, we calculate SSIM and DTWD, for the 50 test samples. The results for all metrics are presented in Table 3.

---
[11]https://huggingface.co/microsoft/wavlm-base-plus-sv
[12]Please note that directly training the GPT-Talker model on the DailyTalk dataset leads to poor performance. Therefore, the experimental results in this section are obtained by fine-tuning all systems based on the pre-training described in section 3.4.

Table 3: Subjective and objective experimental results on DailyTalk and NCSSD datasets (CL-EN & RC-EN).

| Methods | DailyTalk | | | | NCSSD (EN) | | | |
|---|---|---|---|---|---|---|---|---|
| | SSIM (↑) | DTWD (↓) | N-DMOS (↑) | E-DMOS (↑) | SSIM (↑) | DTWD (↓) | N-DMOS (↑) | E-DMOS (↑) |
| CCATTS [17] | 0.734 | 67.376 | 3.402 ± 0.025 | 3.429 ± 0.019 | 0.752 | 65.234 | 3.425 ±0.025 | 3.493 ± 0.022 |
| FCTalker [19] | 0.741 | 65.241 | 3.405 ± 0.026 | 3.537 ± 0.016 | 0.756 | 65.375 | 3.490 ± 0.029 | 3.491 ± 0.023 |
| ECSS [34] | 0.749 | 64.564 | 3.597 ± 0.024 | 3.585 ± 0.028 | 0.761 | 63.654 | 3.507 ± 0.018 | 3.587 ± 0.031 |
| **GPT-Talker (Ours)** | **0.882** | **42.125** | **3.890 ± 0.033** | **3.908 ± 0.029** | **0.884** | **45.627** | **3.884 ± 0.031** | **3.891 ± 0.017** |
| **Ground Truth** | - | - | 4.486 ± 0.026 | 4.501 ± 0.025 | - | - | 4.399 ± 0.020 | 4.493 ± 0.027 |

Table 4: Subjective and objective experimental results on the analysis of three-stage training. (∗ means suboptimal result.)

| Training Strategy | NCSSD (EN) | | | | NCSSD (ZH) | | | |
|---|---|---|---|---|---|---|---|---|
| | SSIM (↑) | DTWD (↓) | N-DMOS (↑) | E-DMOS (↑) | SSIM (↑) | DTWD (↓) | N-DMOS (↑) | E-DMOS (↑) |
| One-Stage (w/ CL & RC) | 0.843 | 57.734 | 3.609 ± 0.024 | 3.698 ± 0.014 | 0.851 | 56.842 | 3.625 ± 0.026 | 3.731 ± 0.021 |
| Two-Stage (w/ CL) | 0.875 | 48.653 | 3.713 ± 0.021 | 3.714 ± 0.017 | 0.877 | 47.854 | 3.716 ± 0.018 | 3.690 ± 0.030 |
| Two-Stage (w/ RC) | 0.879 | 47.863 | 3.716 ± 0.027 | 3.781 ± 0.023 | 0.884 | 48.125 | 3.696 ± 0.018 | 3.709 ± 0.029 |
| Two-Stage (w/ CL & RC) | 0.888 | 44.834 | 3.902 ± 0.033 | 3.925 ± 0.028 | 0.891 | 44.682 | 3.910 ± 0.030 | 3.916 ± 0.032 |
| **Three-Stage (Ours)** | **0.904** | **42.076** | **3.910 ± 0.019** | 3.922 ± 0.022∗ | **0.908** | **43.002** | 3.906 ± 0.020∗ | **3.987 ± 0.021** |

From the DTWD results in the table, it is evident that the speech synthesized by GPT-Talker outperforms the baselines, indicating that GPT-Talker is more capable of capturing pitch-related expressiveness similar to the GT. Furthermore, the N-DMOS and E-DMOS results show that our GPT-Talker significantly outperforms all the baselines. For example, in the DailyTalk dataset, GPT-Talker achieves N-DMOS and E-DMOS scores of 3.890 and 3.908, respectively, while the baselines score below 3.6. Similar trends are observed in the results for the NCSSD dataset. Moreover, GPT-Talker exhibits closer resemblance to GT compared to the baselines. These results demonstrate that GPT-Talker, leveraging ConGPT, effectively models the impact of context on the semantics and style of the current sentence, while ConVITS successfully renders semantic and stylistic aspects in CSS, resulting in highly natural and expressive synthesized speech. Lastly, by observing the SSIM metric results, it can be seen that GPT-Talker achieves the highest speaker similarity compared to all the baselines, proving that our ConVITS excels in timbre rendering on top of semantic and stylistic rendering. With the dialogue context modeling ability of ConGPT, ConVITS achieves highly expressive conversational speech synthesis.

### 6.3 Analysis of Three-Stage Training

This section validates the three-stage strategy, introduced in Section 3.4, used to train GPT-Talker. We design the following four training strategies for validation: 1) **One-Stage (w/ CL&RC)** means we directly use both the collection and recording subsets of NCSSD to train the GPT-Talker; 2) **Two-Stage (w/ CL&RC)** means we follow the same method as described in section 3.4 for pre-training. Afterward, we merge the two subsets together and perform fine-tuning; 3) **Two-Stage (w/ CL)**. This method is similar to the second one, with the difference that during fine-tuning, only the collection subset is selected; 4) **Two-Stage (w/ RC)**: This method is similar to the second one, with the difference that during fine-tuning, only the recording subset is selected. We conduct subjective and objective experiments using the same configuration as in the previous section and report all results in Table 4. We observe that the multi-stage approach outperforms the single-stage approach in all metrics, suggesting that using a pre-training strategy enables the model to learn basic speech generation capabilities, and fine-tuning on the conversational data helps improve dialogue context understanding and expression abilities. Additionally, the "Two-Stage (w/ CL&RC)" and our three-stage approaches outperform previous two-stage methods, demonstrating that fine-tuning on the entire conversational dataset allows the model to learn dialogue expression capabilities with the support of a large volume of data. Comparing the "Two-Stage (w/ CL&RC)" approach with our three-stage approach, although there is a slight lag in one metric for both the Chinese and English datasets, our three-stage approach exhibits significant advantages in the remaining six metrics. This confirms that our three-stage method gradually guides the model to learn richer dialogue expression information, leading to better conversational speech generation capabilities.

## 7 Conclusion

In this work, we propose a novel GPT-based conversational speech synthesis (CSS) model, termed GPT-Talker, for user-agent interaction. It consists of ConGPT and ConVITS to model the semantic and style expression in the unified user-agent discrete dialogue context sequence and infer the expressive speech for the agent. We also propose the largest-scale conversational speech synthesis dataset to date, termed NCSSD, which stands out in terms of language and data source diversity. This dataset can support the training of GPT-Talker and even future GPT-style CSS models, providing a valuable resource for advancing CSS technology. The comprehensive experiments are conducted on the reliability of the NCSSD and the effectiveness of our GPT-Talker. We encourage the research community to use the NCSSD for spoken dialogue modeling.

## 8 Acknowledgments

The research by Rui Liu was funded by the Young Scientists Fund of the National Natural Science Foundation of China (No. 62206136), and Guangdong Provincial Key Laboratory of Human Digital Twin (No. 2022B1212010004), and the "Inner Mongolia Science and Technology Achievement Transfer and Transformation Demonstration Zone, University Collaborative Innovation Base, and University Entrepreneurship Training Base" Construction Project (Supercomputing Power Project) (No.21300-231510). The research by Yifan Hu was funded by the Inner Mongolia University 2024 Graduate Student Research and Innovation Project (Key Project Grant No. 11200-5223737). The research by Haizhou Li was funded by National Natural Science Foundation of China (Grant No. 62271432), Shenzhen Science and Technology Program ZDSYS20230626091302006, and Shenzhen Science and Technology Research Fund (Fundamental Research Key Project Grant No. JCYJ20220818103001002).

## References

[1] Zahra Abbasiantaeb, Yifei Yuan, Evangelos Kanoulas, and Mohammad Aliannejadi. 2024. Let the llms talk: Simulating human-to-human conversational qa via zero-shot llm-to-llm interactions. In *Proceedings of the 17th ACM International Conference on Web Search and Data Mining*. 8–17.

[2] Tom B. Brown, Benjamin Mann, Nick Ryder, Melanie Subbiah, Jared Kaplan, Prafulla Dhariwal, Arvind Neelakantan, Pranav Shyam, Girish Sastry, Amanda Askell, Sandhini Agarwal, Ariel Herbert-Voss, Gretchen Krueger, Tom Henighan, Rewon Child, Aditya Ramesh, Daniel M. Ziegler, Jeffrey Wu, Clemens Winter, Christopher Hesse, Mark Chen, Eric Sigler, Mateusz Litwin, Scott Gray, Benjamin Chess, Jack Clark, Christopher Berner, Sam McCandlish, Alec Radford, Ilya Sutskever, and Dario Amodei. 2020. Language Models are Few-Shot Learners. In *Advances in Neural Information Processing Systems 33: Annual Conference on Neural Information Processing Systems 2020, NeurIPS 2020, December 6-12, 2020, virtual*, Hugo Larochelle, Marc'Aurelio Ranzato, Raia Hadsell, Maria-Florina Balcan, and Hsuan-Tien Lin (Eds.). https://proceedings.neurips.cc/paper/2020/hash/1457c0d6bfcb4967418bfb8ac142f64a-Abstract.html

[3] Paweł Budzianowski, Taras Sereda, Tomasz Cichy, and Ivan Vulić. 2024. Pheme: Efficient and Conversational Speech Generation. *arXiv preprint arXiv:2401.02839* (2024).

[4] Carlos Busso, Murtaza Bulut, Chi-Chun Lee, Abe Kazemzadeh, Emily Mower, Samuel Kim, Jeannette N. Chang, Sungbok Lee, and Shrikanth S. Narayanan. 2008. IEMOCAP: interactive emotional dyadic motion capture database. *Lang. Resour. Evaluation* 42, 4 (2008), 335–359. https://doi.org/10.1007/S10579-008-9076-6

[5] Santiago Castro, Devamanyu Hazarika, Verónica Pérez-Rosas, Roger Zimmermann, Rada Mihalcea, and Soujanya Poria. 2019. Towards Multimodal Sarcasm Detection (An _Obviously_ Perfect Paper). In *Proceedings of the 57th Annual Meeting of the Association for Computational Linguistics*. 4619–4629.

[6] Bei Chen, Fengji Zhang, Anh Nguyen, Daoguang Zan, Zeqi Lin, Jian-Guang Lou, and Weizhu Chen. 2023. CodeT: Code Generation with Generated Tests. In *The Eleventh International Conference on Learning Representations, ICLR 2023, Kigali, Rwanda, May 1-5, 2023*. OpenReview.net. https://openreview.net/pdf?id=ktrw68Cmu9c

[7] Li-Wei Chen, Shinji Watanabe, and Alexander Rudnicky. 2023. A vector quantized approach for text to speech synthesis on real-world spontaneous speech. In *Proceedings of the AAAI Conference on Artificial Intelligence*, Vol. 37. 12644–12652.

[8] Yirong Chen, Weiquan Fan, Xiaofen Xing, Jianxin Pang, Minlie Huang, Wenjing Han, Qianfeng Tie, and Xiangmin Xu. 2022. CPED: A Large-Scale Chinese Personalized and Emotional Dialogue Dataset for Conversational AI. *CoRR* abs/2205.14727 (2022). https://doi.org/10.48550/ARXIV.2205.14727 arXiv:2205.14727

[9] Yayue Deng, Jinlong Xue, Yukang Jia, Qifei Li, Yichen Han, Fengping Wang, Yingming Gao, Dengfeng Ke, and Ya Li. 2023. CONCSS: Contrastive-based Context Comprehension for Dialogue-appropriate Prosody in Conversational Speech Synthesis. *CoRR* abs/2312.10358 (2023). https://doi.org/10.48550/ARXIV.2312.10358 arXiv:2312.10358

[10] Yayue Deng, Jinlong Xue, Fengping Wang, Yingming Gao, and Ya Li. 2023. CMCU-CSS: Enhancing Naturalness via Commonsense-based Multi-modal Context Understanding in Conversational Speech Synthesis. In *Proceedings of the 31st ACM International Conference on Multimedia, MM 2023, Ottawa, ON, Canada, 29 October 2023- 3 November 2023*, Abdulmotaleb El-Saddik, Tao Mei, Rita Cucchiara, Marco Bertini, Diana Patricia Tobon Vallejo, Pradeep K. Atrey, and M. Shamim Hossain (Eds.). ACM, 6081–6089. https://doi.org/10.1145/3581783.3612565

[11] Yayue Deng, Jinlong Xue, Fengping Wang, Yingming Gao, and Ya Li. 2023. CMCU-CSS: Enhancing Naturalness via Commonsense-based Multi-modal Context Understanding in Conversational Speech Synthesis. In *Proceedings of the 31st ACM International Conference on Multimedia, MM 2023, Ottawa, ON, Canada, 29 October 2023- 3 November 2023*, Abdulmotaleb El-Saddik, Tao Mei, Rita Cucchiara, Marco Bertini, Diana Patricia Tobon Vallejo, Pradeep K. Atrey, and M. Shamim Hossain (Eds.). ACM, 6081–6089. https://doi.org/10.1145/3581783.3612565

[12] Pranay Dighe, Yi Su, Shangshang Zheng, Yunshu Liu, Vineet Garg, Xiaochuan Niu, and Ahmed Tewfik. 2024. Leveraging large language models for exploiting asr uncertainty. In *ICASSP 2024-2024 IEEE International Conference on Acoustics, Speech and Signal Processing (ICASSP)*. IEEE, 12231–12235.

[13] Yasser El Miedany and Yasser El Miedany. 2019. Virtual reality and augmented reality. *Rheumatology teaching: the art and science of medical education* (2019), 403–427.

[14] Luciano Floridi and Massimo Chiriatti. 2020. GPT-3: Its nature, scope, limits, and consequences. *Minds and Machines* 30 (2020), 681–694.

[15] ARCHISMITA GHOSH, GADDAM PRATHIK KUMAR, PAARTH PRASAD, DHEERAJ KUMAR, SAMYAK JAIN, and JATIN CHOPRA. 2023. Synergizing Generative Intelligence: Advancements in Artificial Intelligence for Intelligent Vehicle Systems and Vehicular Networks. (2023).

[16] GPT-SoVITS. 2024. https://github.com/RVC-Boss/GPT-SoVITS.

[17] Haohan Guo, Shaofei Zhang, Frank K Soong, Lei He, and Lei Xie. 2021. Conversational end-to-end tts for voice agents. In *2021 IEEE Spoken Language Technology Workshop (SLT)*. IEEE, 403–409.

[18] Wei-Ning Hsu, Benjamin Bolte, Yao-Hung Hubert Tsai, Kushal Lakhotia, Ruslan Salakhutdinov, and Abdelrahman Mohamed. 2021. HuBERT: Self-Supervised Speech Representation Learning by Masked Prediction of Hidden Units. *IEEE ACM Trans. Audio Speech Lang. Process.* 29 (2021), 3451–3460. https://doi.org/10.1109/TASLP.2021.3122291

[19] Yifan Hu, Rui Liu, Guanglai Gao, and Haizhou Li. 2022. FCTalker: Fine and Coarse Grained Context Modeling for Expressive Conversational Speech Synthesis. *CoRR* abs/2210.15360 (2022). https://doi.org/10.48550/ARXIV.2210.15360 arXiv:2210.15360

[20] Keith Ito and Linda Johnson. 2017. The lj speech dataset. 2017. *URL https://keithito.com/LJ-Speech-Dataset* (2017).

[21] Mahyuzie Jenal, Athira Nabilla Omar, Muhammad Azizi Aswad Hisham, Wan Najmi Wan Mohd Noh, and Zul Adib Izzuddin Razali. 2022. Smart home controlling system. *Journal of Electronic Voltage and Application* 3, 1 (2022), 92–104.

[22] Ziyue Jiang, Jinglin Liu, Yi Ren, Jinzheng He, Chen Zhang, Zhenhui Ye, Pengfei Wei, Chunfeng Wang, Xiang Yin, Zejun Ma, and Zhou Zhao. 2023. Mega-TTS 2: Zero-Shot Text-to-Speech with Arbitrary Length Speech Prompts. *CoRR* abs/2307.07218 (2023). https://doi.org/10.48550/ARXIV.2307.07218 arXiv:2307.07218

[23] Heeseung Kim, Soonshin Seo, Kyeongseok Jeong, Ohsung Kwon, Jungwhan Kim, Jaehong Lee, Eunwoo Song, Myungwoo Oh, Sungroh Yoon, and Kang Min Yoo. 2024. Unified Speech-Text Pretraining for Spoken Dialog Modeling. *arXiv preprint arXiv:2402.05706* (2024).

[24] Heeseung Kim, Soonshin Seo, Kyeongseok Jeong, Ohsung Kwon, Jungwhan Kim, Jaehong Lee, Eunwoo Song, Myungwoo Oh, Sungroh Yoon, and Kang Min Yoo. 2024. Unified Speech-Text Pretraining for Spoken Dialog Modeling. *CoRR* abs/2402.05706 (2024). https://doi.org/10.48550/ARXIV.2402.05706 arXiv:2402.05706

[25] Jaehyeon Kim, Jungil Kong, and Juhee Son. 2021. Conditional variational autoencoder with adversarial learning for end-to-end text-to-speech. In *International Conference on Machine Learning*. PMLR, 5530–5540.

[26] Kushal Lakhotia, Eugene Kharitonov, Wei-Ning Hsu, Yossi Adi, Adam Polyak, Benjamin Bolte, Tu-Anh Nguyen, Jade Copet, Alexei Baevski, Abdelrahman Mohamed, et al. 2021. On generative spoken language modeling from raw audio. *Transactions of the Association for Computational Linguistics* 9 (2021), 1336–1354.

[27] Keon Lee, Kyumin Park, and Daeyoung Kim. 2023. DailyTalk: Spoken Dialogue Dataset for Conversational Text-to-Speech. In *IEEE International Conference on Acoustics, Speech and Signal Processing ICASSP 2023, Rhodes Island, Greece, June 4-10, 2023*. IEEE, 1–5. https://doi.org/10.1109/ICASSP49357.2023.10095751

[28] Jingbei Li, Yi Meng, Chenyi Li, Zhiyong Wu, Helen Meng, Chao Weng, and Dan Su. 2022. Enhancing Speaking Styles in Conversational Text-to-Speech Synthesis with Graph-Based Multi-Modal Context Modeling. In *IEEE International Conference on Acoustics, Speech and Signal Processing, ICASSP 2022, Virtual and Singapore, 23-27 May 2022*. IEEE, 7917–7921. https://doi.org/10.1109/ICASSP43922.2022.9747837

[29] Jingbei Li, Yi Meng, Xixin Wu, Zhiyong Wu, Jia Jia, Helen Meng, Qiao Tian, Yuping Wang, and Yuxuan Wang. 2022. Inferring speaking styles from multi-modal conversational context by multi-scale relational graph convolutional networks. In *Proceedings of the 30th ACM International Conference on Multimedia*. 5811–5820.

[30] Yanran Li, Hui Su, Xiaoyu Shen, Wenjie Li, Ziqiang Cao, and Shuzi Niu. 2017. DailyDialog: A Manually Labelled Multi-turn Dialogue Dataset. In *Proceedings of the Eighth International Joint Conference on Natural Language Processing, IJCNLP 2017, Taipei, Taiwan, November 27 - December 1, 2017 - Volume 1: Long Papers*, Greg Kondrak and Taro Watanabe (Eds.). Asian Federation of Natural Language Processing, 986–995. https://aclanthology.org/I17-1099/

[31] Guan-Ting Lin, Prashanth Gurunath Shivakumar, Ankur Gandhe, Chao-Han Huck Yang, Yile Gu, Shalini Ghosh, Andreas Stolcke, Hung-yi Lee, and Ivan Bulyko. 2023. Paralinguistics-Enhanced Large Language Modeling of Spoken Dialogue. *CoRR* abs/2312.15316 (2023). https://doi.org/10.48550/ARXIV.2312.15316 arXiv:2312.15316

[32] Guan-Ting Lin, Cheng-Han Chiang, and Hung-yi Lee. 2024. Advancing Large Language Models to Capture Varied Speaking Styles and Respond Properly in Spoken Conversations. *arXiv preprint arXiv:2402.12786* (2024).

[33] Guan-Ting Lin, Prashanth Gurunath Shivakumar, Ankur Gandhe, Chao-Han Huck Yang, Yile Gu, Shalini Ghosh, Andreas Stolcke, Hung-yi Lee, and Ivan Bulyko. 2024. Paralinguistics-enhanced large language modeling of spoken dialogue. In *ICASSP 2024-2024 IEEE International Conference on Acoustics, Speech and Signal Processing (ICASSP)*. IEEE, 10316–10320.

[34] Rui Liu, Yifan Hu, Yi Ren, Xiang Yin, and Haizhou Li. 2024. Emotion Rendering for Conversational Speech Synthesis with Heterogeneous Graph-Based Context Modeling. *Proceedings of the AAAI Conference on Artificial Intelligence* 38, 17 (Mar. 2024), 18698–18706. https://doi.org/10.1609/aaai.v38i17.29833

[35] Yuchen Liu, Haoyu Zhang, Shichao Liu, Xiang Yin, Zejun Ma, and Qin Jin. 2023. Emotionally Situated Text-to-Speech Synthesis in User-Agent Conversation. In *Proceedings of the 31st ACM International Conference on Multimedia, MM 2023, Ottawa, ON, Canada, 29 October 2023- 3 November 2023*, Abdulmotaleb El-Saddik, Tao Mei, Rita Cucchiara, Marco Bertini, Diana Patricia Tobon Vallejo, Pradeep K. Atrey, and M. Shamim Hossain (Eds.). ACM, 5966–5974. https://doi.org/10.1145/3581783.3613823

[36] Yasmin Moslem, Rejwanul Haque, John D. Kelleher, and Andy Way. 2023. Adaptive Machine Translation with Large Language Models. In *Proceedings of the 24th Annual Conference of the European Association for Machine Translation, EAMT 2023, Tampere, Finland, 12-15 June 2023*, Mary Nurminen, Judith Brenner, Maarit Koponen, Sirkku Latomaa, Mikhail Mikhailov, Frederike Schierl, Tharindu Ranasinghe, Eva Vanmassenhove, Sergi Alvarez Vidal, Nora Aranberri, Mara Nunziatini, Carla Parra Escartín, Mikel L. Forcada, Maja Popovic, Carolina Scarton, and Helena Moniz (Eds.). European Association for Machine Translation, 227–237. https://aclanthology.org/2023.eamt-1.22

[37] Meinard Müller. 2007. Dynamic time warping. *Information retrieval for music and motion* (2007), 69–84.

[38] Tu Anh Nguyen, Eugene Kharitonov, Jade Copet, Yossi Adi, Wei-Ning Hsu, Ali Elkahky, Paden Tomasello, Robin Algayres, Benoît Sagot, Abdelrahman Mohamed, and Emmanuel Dupoux. 2022. Generative Spoken Dialogue Language Modeling. *CoRR* abs/2203.16502 (2022). https://doi.org/10.48550/ARXIV.2203.16502 arXiv:2203.16502

[39] Yuto Nishimura, Yuki Saito, Shinnosuke Takamichi, Kentaro Tachibana, and Hiroshi Saruwatari. 2022. Acoustic Modeling for End-to-End Empathetic Dialogue Speech Synthesis Using Linguistic and Prosodic Contexts of Dialogue History. In *Proc. Interspeech 2022*. 3373–3377. https://doi.org/10.21437/Interspeech.2022-403

[40] Pilar Oplustil-Gallegos, Johannah O'Mahony, and Simon King. 2021. Comparing acoustic and textual representations of previous linguistic context for improving Text-to-Speech. In *Proc. 11th ISCA Speech Synthesis Workshop (SSW 11)*. 205–210. https://doi.org/10.21437/SSW.2021-36

[41] Soujanya Poria, Devamanyu Hazarika, Navonil Majumder, Gautam Naik, Erik Cambria, and Rada Mihalcea. 2019. MELD: A Multimodal Multi-Party Dataset for Emotion Recognition in Conversations. In *Proceedings of the 57th Conference of the Association for Computational Linguistics, ACL 2019, Florence, Italy, July 28- August 2, 2019, Volume 1: Long Papers*, Anna Korhonen, David R. Traum, and Lluís Màrquez (Eds.). Association for Computational Linguistics, 527–536. https://doi.org/10.18653/V1/P19-1050

[42] Yi Ren, Chenxu Hu, Xu Tan, Tao Qin, Sheng Zhao, Zhou Zhao, and Tie-Yan Liu. 2020. FastSpeech 2: Fast and High-Quality End-to-End Text to Speech. In *International Conference on Learning Representations*.

[43] Yuki Saito, Eiji Iimori, Shinnosuke Takamichi, Kentaro Tachibana, and Hiroshi Saruwatari. 2023. CALLS: Japanese Empathetic Dialogue Speech Corpus of Complaint Handling and Attentive Listening in Customer Center. *CoRR* abs/2305.13713 (2023). https://doi.org/10.48550/ARXIV.2305.13713 arXiv:2305.13713

[44] Yuki Saito, Yuto Nishimura, Shinnosuke Takamichi, Kentaro Tachibana, and Hiroshi Saruwatari. 2022. STUDIES: Corpus of Japanese Empathetic Dialogue Speech Towards Friendly Voice Agent. In *Interspeech 2022, 23rd Annual Conference of the International Speech Communication Association, Incheon, Korea, 18-22 September 2022*, Hanseok Ko and John H. L. Hansen (Eds.). ISCA, 5155–5159. https://doi.org/10.21437/INTERSPEECH.2022-300

[45] Yao Shi, Hui Bu, Xin Xu, Shaoji Zhang, and Ming Li. 2020. AISHELL-3: A Multi-speaker Mandarin TTS Corpus and the Baselines. *CoRR* abs/2010.11567 (2020). arXiv:2010.11567 https://arxiv.org/abs/2010.11567

[46] Ben Swanson, Kory W. Mathewson, Ben Pietrzak, Sherol Chen, and Monica Dinalescu. 2021. Story Centaur: Large Language Model Few Shot Learning as a Creative Writing Tool. In *Proceedings of the 16th Conference of the European Chapter of the Association for Computational Linguistics: System Demonstrations, EACL 2021, Online, April 19-23, 2021*, Dimitra Gkatzia and Djamé Seddah (Eds.). Association for Computational Linguistics, 244–256. https://doi.org/10.18653/V1/2021.EACL-DEMOS.29

[47] Minh Duc Vu, Han Wang, Zhuang Li, Jieshan Chen, Shengdong Zhao, Zhenchang Xing, and Chunyang Chen. 2024. GPTVoiceTasker: LLM-Powered Virtual Assistant for Smartphone. *arXiv preprint arXiv:2401.14268* (2024).

[48] Chengyi Wang, Sanyuan Chen, Yu Wu, Ziqiang Zhang, Long Zhou, Shujie Liu, Zhuo Chen, Yanqing Liu, Huaming Wang, Jinyu Li, Lei He, Sheng Zhao, and Furu Wei. 2023. Neural Codec Language Models are Zero-Shot Text to Speech Synthesizers. *CoRR* abs/2301.02111 (2023). https://doi.org/10.48550/ARXIV.2301.02111 arXiv:2301.02111

[49] Jiaming Wang, Zhihao Du, Qian Chen, Yunfei Chu, Zhifu Gao, Zerui Li, Kai Hu, Xiaohuan Zhou, Jin Xu, Ziyang Ma, Wen Wang, Siqi Zheng, Chang Zhou, Zhijie Yan, and Shiliang Zhang. 2023. LauraGPT: Listen, Attend, Understand, and Regenerate Audio with GPT. *CoRR* abs/2310.04673 (2023). https://doi.org/10.48550/ARXIV.2310.04673 arXiv:2310.04673

[50] Anuradha Welivita, Yubo Xie, and Pearl Pu. 2020. Fine-grained Emotion and Intent Learning in Movie Dialogues. *CoRR* abs/2012.13624 (2020). arXiv:2012.13624 https://arxiv.org/abs/2012.13624

[51] Jian Wu, Yashesh Gaur, Zhuo Chen, Long Zhou, Yimeng Zhu, Tianrui Wang, Jinyu Li, Shujie Liu, Bo Ren, Linquan Liu, et al. 2023. On decoder-only architecture for speech-to-text and large language model integration. In *2023 IEEE Automatic Speech Recognition and Understanding Workshop (ASRU)*. IEEE, 1–8.

[52] Hongfei Xue, Yuhao Liang, Bingshen Mu, Shiliang Zhang, Mengzhe Chen, Qian Chen, and Lei Xie. 2024. E-chat: Emotion-sensitive Spoken Dialogue System with Large Language Models. *CoRR* abs/2401.00475 (2024). https://doi.org/10.48550/ARXIV.2401.00475 arXiv:2401.00475

[53] Jinlong Xue, Yayue Deng, Fengping Wang, Ya Li, Yingming Gao, Jianhua Tao, Jianqing Sun, and Jiaen Liang. 2023. M 2-ctts: End-to-end multi-scale multi-modal conversational text-to-speech synthesis. In *ICASSP 2023-2023 IEEE International Conference on Acoustics, Speech and Signal Processing (ICASSP)*. IEEE, 1–5.

[54] Rohola Zandie, Mohammad H. Mahoor, Julia Madsen, and Eshrat S. Emamian. 2021. RyanSpeech: A Corpus for Conversational Text-to-Speech Synthesis. In *Interspeech 2021, 22nd Annual Conference of the International Speech Communication Association, Brno, Czechia, 30 August - 3 September 2021*, Hynek Hermansky, Honza Cernocký, Lukás Burget, Lori Lamel, Odette Scharenborg, and Petr Motlícek (Eds.). ISCA, 2751–2755. https://doi.org/10.21437/INTERSPEECH.2021-341

[55] Heiga Zen, Viet Dang, Rob Clark, Yu Zhang, Ron J. Weiss, Ye Jia, Zhifeng Chen, and Yonghui Wu. 2019. LibriTTS: A Corpus Derived from LibriSpeech for Text-to-Speech. In *Interspeech 2019, 20th Annual Conference of the International Speech Communication Association, Graz, Austria, 15-19 September 2019*, Gernot Kubin and Zdravko Kacic (Eds.). ISCA, 1526–1530. https://doi.org/10.21437/INTERSPEECH.2019-2441

[56] Dong Zhang, Shimin Li, Xin Zhang, Jun Zhan, Pengyu Wang, Yaqian Zhou, and Xipeng Qiu. 2023. SpeechGPT: Empowering Large Language Models with Intrinsic Cross-Modal Conversational Abilities. In *Findings of the Association for Computational Linguistics: EMNLP 2023*. 15757–15773.

[57] Dong Zhang, Shimin Li, Xin Zhang, Jun Zhan, Pengyu Wang, Yaqian Zhou, and Xipeng Qiu. 2023. SpeechGPT: Empowering Large Language Models with Intrinsic Cross-Modal Conversational Abilities. In *Findings of the Association for Computational Linguistics: EMNLP 2023, Singapore, December 6-10, 2023*, Houda Bouamor, Juan Pino, and Kalika Bali (Eds.). Association for Computational Linguistics, 15757–15773. https://aclanthology.org/2023.findings-emnlp.1055

[58] Yizhe Zhang, Siqi Sun, Michel Galley, Yen-Chun Chen, Chris Brockett, Xiang Gao, Jianfeng Gao, Jingjing Liu, and Bill Dolan. 2020. DIALOGPT : Large-Scale Generative Pre-training for Conversational Response Generation. In *Proceedings of the 58th Annual Meeting of the Association for Computational Linguistics: System Demonstrations, ACL 2020, Online, July 5-10, 2020*, Asli Celikyilmaz and Tsung-Hsien Wen (Eds.). Association for Computational Linguistics, 270–278. https://doi.org/10.18653/V1/2020.ACL-DEMOS.30

[59] Jinming Zhao, Tenggan Zhang, Jingwen Hu, Yuchen Liu, Qin Jin, Xinchao Wang, and Haizhou Li. 2022. M3ED: Multi-modal Multi-scene Multi-label Emotional Dialogue Database. In *Proceedings of the 60th Annual Meeting of the Association for Computational Linguistics (Volume 1: Long Papers), ACL 2022, Dublin, Ireland, May 22-27, 2022*, Smaranda Muresan, Preslav Nakov, and Aline Villavicencio (Eds.). Association for Computational Linguistics, 5699–5710. https://doi.org/10.18653/V1/2022.ACL-LONG.391

# Supplementary Materials: Generative Expressive Conversational Speech Synthesis

## A MORE INFORMATION ABOUT THE NCSSD

### A.1 The Detailed Data Source

The collection subset of NCSSD, CL-EN and CL-ZH, are extracted from 79 English and 34 Chinese TV shows respectively. Please refer to the entire TV show list in Figure 1 (The Roman numerals from I to X represent the seasons of a specfic TV show).

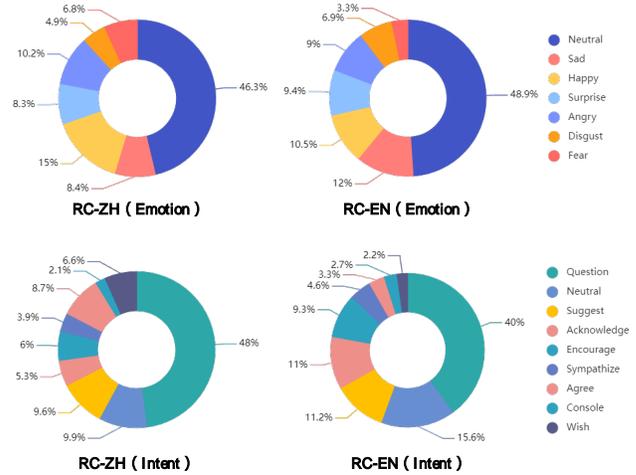

Figure 2: Statistics of emotion and intent labels in the recorded data of NCSSD (RC-EN and RC-ZH).

Table 1: Detailed splits for each subset of NCSSD.

| SubSet | Language | Dialogues | | | |
|---|---|---|---|---|---|
| | | Train | Valid | Test | All |
| CL | EN | 5626 | 707 | 700 | 7033 |
| | ZH | 7120 | 906 | 750 | 8776 |
| RC | EN | 957 | 139 | 100 | 1196 |
| | ZH | 2006 | 245 | 200 | 2451 |

### A.3 More Data Statistics

*A.3.1 Emotion and Intent statistics.* During the development of the recording subsets for NCSSD (RC-EN and RC-ZH), we employ ChatGPT to create scripts for two-person dialogues, along with emotional and intent labels for each sentence to help the recording volunteers. As show in Figure 2, we separately statistic the distribution of seven emotion labels and nine intent labels in RC-EN and RC-ZH.

*A.3.2 NCSSD splits.* Table 1 presents the division of Train, Valid, and Test sets for the number of dialogues in each subset of NCSSD.

### A.4 Ethics Statement

This work provides the NCSSD dataset to the research community for free, which is used for conversational speech synthesis research.

For the collection subset (CL-EN and CL-ZH), which primarily derives from Chinese and English TV shows, we employ two Chinese college students to collect 79 TV shows and use our automatic pipeline to extract high-quality dialogues. To ensure equitable compensation, we pay them 50 yuan per hour ($6.91 USD), which is considered fair locally. Due to copyright issues with TV shows, we

Figure 1: Detailed list of TV Shows for collection subset of NCSSD.

### A.2 The Construction Workflow of NCSSD

As shown in Figure 3, we visualize the construction process of NCSSD. The upper panel displays the automatic pipeline for collection subset, including Video Selection, Dialogue Scene Extraction, Dialogue Segment Extraction, and Dialogue Script Recognition. The lower panel shows the workflow for recording subset, including Dialogue Script Draft Generation, Spoken Dialogue Recording, and Dialogue Script Re-identification.

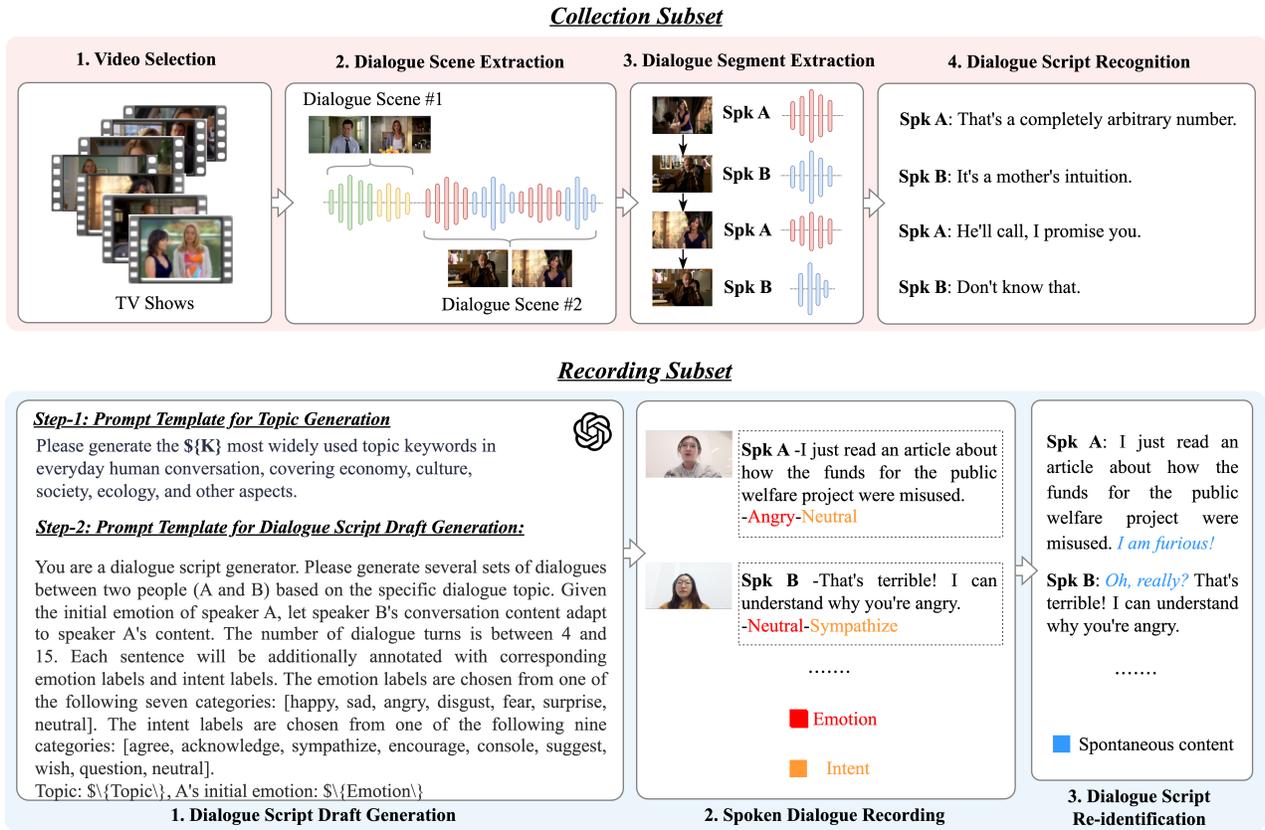

Figure 3: The construction process of NCSSD. The upper panel shows the automatic pipeline of the collection subset, while the bottom panel shows the workflow of the recording subset.

will only disclose the names of the TV shows, dialogue scripts, and acoustic features, excluding visual features in the dialogue. Certainly, other researchers can also input videos of other TV shows into our automatic pipeline to extract the dialogue segments.

For the recording subset (RC-EN and RC-ZH), we employ 15 female and 12 male young adults for on-site recording, with a compensation of 800 yuan ($110.48 USD) per hour. We sign a public usage agreement with each recording personnel, planning to publicly release the transcribed scripts, acoustic features, and facial visual features of the recorded dialogues.

We believe that these high-quality multimodal dialogue resources advance research in conversational speech synthesis.

## B EXPERIMENT SETUP

In ConGPT module, the phoneme encoder utilizes existing method g2p_en[1] and opencpop-strict dictionary [2] for text discretization. The unit encoder includes a 1D convolutional layer with a stride of 2, and a vector quantization module with a dimension of 256 and a codebook size of 1024. ConGPT features 24 Transformer encoder blocks, each with a Feed Forward block dimension of 2048, 16 attention heads, and a dropout rate of 0.1. During training, the number of dialogue turns, $N$, is set to 3. For the ConVITS module, the Content Encoder includes a text embedding layer with an output dimension of 192 and 6 transformer encoder blocks, also with an output dimension of 192; the Token Encoder consists of three transformer encoder blocks with an output dimension of 192. These blocks have a dropout rate of 0.1 and two attention heads. The Timbre Encoder consists of six 2D convolutional blocks with a kernel size of 3 and stride of 2, one GRU layer, and a linear layer with dimensions scaling from 128 to 512. Apart from the duration predictor, the rest of the configurations are the same as the standard VITS.

During the GPT-Talker training process, we used two NVIDIA A100 GPUs and two A800 GPUs, with a batch size set to 8. The NCSSD was divided into four subsets: CL-EN, CL-ZH, RC-EN, RC-ZH, with the number of dialogue groups in the training set, test set, and validation set divided in an 8:1:1 ratio, see the Appendix A for detailed division methods. It is worth noting that our GPT-Talker was trained with 3 turns ($N$) of dialogue, whereas other baselines had 10 turns. To ensure the fairness of the experiments, during inference, we adjusted the dialogue turns of all baselines to 3 turns.

---

[1] https://pypi.org/project/g2p-en/
[2] https://wenet.org.cn/opencpop/

## C MORE RESULTS AND ANALYSIS

### C.1 Emotion Recognition Results

GPT-Talker is trained on a large emotional speech dataset and can generate natural, emotionally rich speech without explicit emotional labels. To demonstrate the emotional expression capabilities of GPT-Talker synthesized speech, we compare it with the state-of-the-art emotional speech synthesis model ECSS and use a pre-trained SER model[3] to predict the emotion categories of 400 audio samples generated by two models, under the DailyTalk and NCSSD (RC-EN) datasets. The correct emotional labels all come from the emotional category labels of the two datasets.

The confusion matrix depicted in the Figure 4 highlights the performance disparities between them. The pronounced diagonal in the matrix confirms that GPT-Talker surpasses ECSS, proving that the emotional conversational speech produced by GPT-Talker exhibits clear emotional expression.

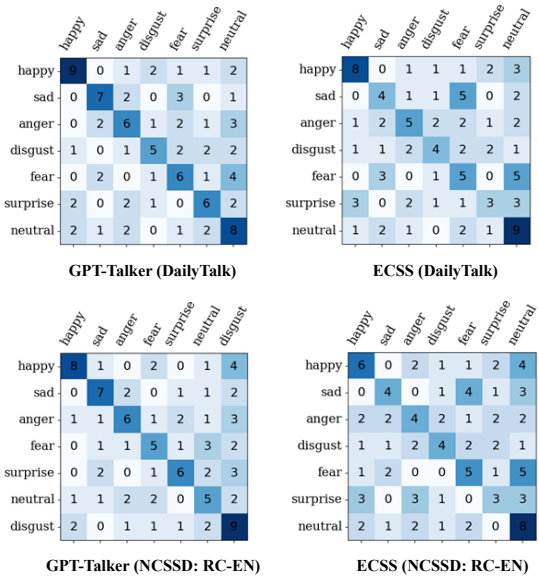

Figure 4: The confusion matrix results of the emotional types of synthetic speech generated by GPT-Talker and ECSS. The X-axis and Y-axis of subfigures represent perceived and true emotion categories.

### C.2 Visualization Analysis (Mel-spectrum and Pitch)

To better understand the quality of GPT-Talker's synthesized speech in terms of naturalness and expressiveness, we analyze the Mel spectrogram and the pitch of its output by comparing it with several baseline models. In example 1 (top of the Figure 5), the corresponding text with synthesized speech from the DailyTalk dataset is "oh, wait a minute. what about a lid for the pan?"; In example 2 (bottom of the Figure 5), the corresponding text with synthesized speech from the NCSSD (EN) dataset is "Great, together we can make a difference and ensure a stronger future for a regime." For example 2, due to limited space for the synthesized speech which is relatively long, we only visualize the first 4 seconds.

---

[3] https://huggingface.co/ehcalabres/wav2vec2-lg-xlsr-en-speech-emotion-recognition

Table 2: Experimental results about various dialogue turns.

| Dialogue Turns ($N$) | DailyTalk | | | |
|---|---|---|---|---|
| | SSIM (↑) | DTWD (↓) | N-DMOS (↑) | E-DMOS (↑) |
| 2 | 0.877 | 52.536 | 3.613+0.027 | 3.636+0.012 |
| 3 | **0.902** | **43.014** | **3.901+0.014** | **3.944+0.023** |
| 4 | 0.713 | 60.382 | 3.324+0.018 | 3.452+0.021 |
| Dialogue Turns ($N$) | NCSSD (EN) | | | |
| | SSIM (↑) | DTWD (↓) | N-DMOS (↑) | E-DMOS (↑) |
| 2 | 0.871 | 53.123 | 3.602+0.011 | 3.620+0.026 |
| 3 | **0.896** | **42.872** | **3.912+0.025** | **3.934+0.016** |
| 4 | 0.732 | 61.621 | 3.331+0.029 | 3.328+0.027 |
| Dialogue Turns ($N$) | NCSSD (ZH) | | | |
| | SSIM (↑) | DTWD (↓) | N-DMOS (↑) | E-DMOS (↑) |
| 2 | 0.883 | 51.764 | 3.636+0.022 | 3.712+0.025 |
| 3 | **0.901** | **43.192** | **3.905+0.015** | **3.956+0.014** |
| 4 | 0.728 | 60.726 | 3.418+0.024 | 3.392+0.035 |

Firstly, the green box in example 1 clearly shows that GPT-Talker exhibits more defined envelope details and superior synthesis quality, as also evidenced in example 2. Secondly, the white polyline in both examples objectively demonstrates that the speech pitch synthesized by GPT-Talker generates richer prosody. Additionally, the red box indicates that the pause frequencies in some of the speech synthesized by GPT-Talker are closer to those of real speech. In a nutshell, our GPT-Talker outperforms all baselines and performs significant advantages in the naturalness and expressiveness of synthesized conversational speech.

### C.3 Dialogue Turns Analysis

To further investigate the influence of dialogue turns (the length of dialogue context) on conversational speech synthesis, we adjust the number of dialogue turns in the input model during inference for comparative analysis. We use models trained on three datasets for validation: DailyTalk, NCSDD (EN) which contains the CL-EN and RC-EN, and NCSDD (ZH) which contains the CL-ZH and RC-ZH. Notably, our model is initially trained with $N$ set at 3, which includes two utterances of dialogue history and one current utterance for synthesis. In this section, we synthesize 50 examples for each length of turns and ask 30 evaluators to conduct subjective evaluations. As evidenced in Table 2, when the dialogue history is reduced to just one sentence ($N$=2), both objective and subjective experimental measures show a decline, suggesting that a more extensive dialogue history enriches the semantic and prosodic style information of the current sentence. However, increasing $N$ to 4, beyond the training configuration of the model, allows for speech synthesis but with noticeably diminished naturalness and emotional expressiveness. This reduction is likely due to the model's constrained ability to handle excessively long sequences. This will be a key issue for our future research.

### C.4 Context Serialization Analysis

In the Con-GPT module of GPT-Talker, the input section utilizes various methods for concatenating dialogue contexts. The first method, which we call $ABAB - Format$, alternates splicing utterances: $\{U_1^t,$

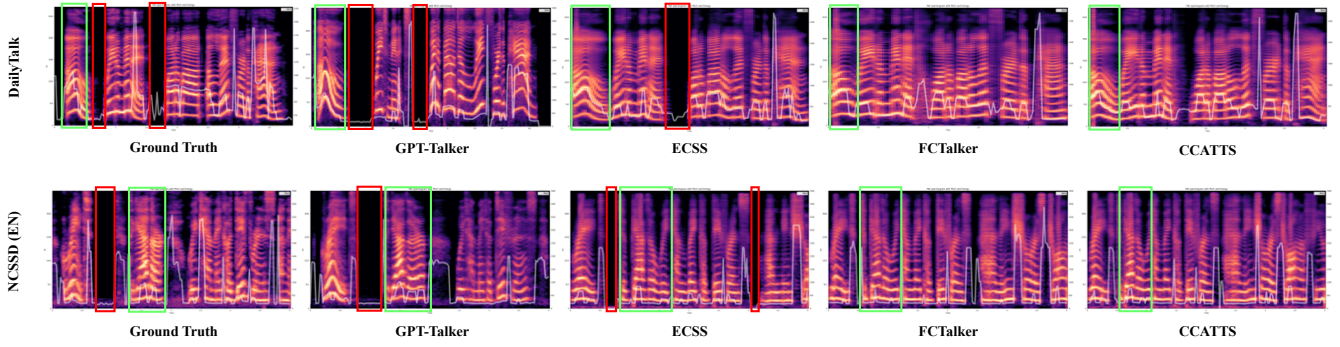

Figure 5: Visualization of Mel-spectrum and Pitch results. The green boxes highlight the spectrum details, while the red boxes highlight the prosody break.

Table 3: The results of different contextual serialization methods comparison.

| Dataset | Method | SSIM (↑) | DTWD (↓) | N-DMOS (↑) | E-DMOS (↑) |
|---|---|---|---|---|---|
| NCSSD (EN) | ABAB-Format | **0.904** | **42.076** | **3.910 ± 0.019** | **3.922 ± 0.022** |
|  | AABB-Format | 0.873 | 44.924 | 3.812 ± 0.015 | 3.861 ± 0.023 |
| NCSSD (ZH) | ABAB-Format | **0.908** | **43.002** | **3.906 ± 0.02** | **3.987 ± 0.021** |
|  | AABB-Format | 0.891 | 45.016 | 3.856 ± 0.033 | 3.877 ± 0.017 |

$U_1^a, U_2^t, U_2^a ..., U_N^t$}, is our chosen approach. The second method, called $AABB - Format$, employs sequential splicing: {$U_1^t, U_2^t ..., U_N^t$, $U_1^a, U_2^a ..., U_{N-1}^a$}. In this section, we use the NCSSD Chinese and English data to train the model. After that, we use each context connection method to synthesize 50 examples, respectively, and ask 30 evaluators to conduct subjective evaluation. According to the experimental findings presented in Table 3, the $ABAB - Format$ method proves more effective for dialogue speech synthesis tasks. This is because this method more closely aligns with the sequence of information exchange in human-machine interactions. Additionally, this method's integration of each utterance's text and audio may help the model more accurately determine the placement of statements, thus fostering a complex interdependency at the sentence level within the dialogue context.

### C.5 Zero-shot Timbre Rendering Performance Analysis

Our model boasts a unique ability to perform zero-shot timbre rendering for the synthesized conversational speech for agent. To showcase this remarkable capability, we feed the model unseen speakers from external datasets, resulting in the successful synthesis of their speech.

Specifically, we randomly sample 20 sets of dialogue speech from 5 unseen speakers in IEMOCAP and M$^3$ED, respectively, and replace one speaker in the original dialogue in DailyTalk and NCSSD (EN and ZH) to synthesize the speech of unseen speakers. We conduct experiments using the speaker similarity evaluation method in 5.3. As the results show in Table 4, the SSIM values are all above 0.82, which is lower than the speaker similarity values within the collection listed in Table 4 of main text, but it also shows that GPT-Talker can basically synthesize the speech of unseen speakers. Of course, a single objective indicator cannot fully reflect the real situation, and we will also place samples on the demo page for readers to reference.

Table 4: Results of zero-shot timbre rendering experiment.

| Evaluation metric | Datasets | | |
|---|---|---|---|
|  | DailyTalk | NCSSD (EN) | NCSSD (ZH) |
| SSIM (↑) | 0.838 | 0.834 | 0.822 |

## D LIMITATIONS AND FUTURE DIRECTIONS

During dataset construction, the recorded and collected dialogues were accompanied by videos, incorporating visual modality information. We haven't explored this aspect in our research yet, so fully utilizing visual modality data to enhance the synthesis of expressive speech will be our subsequent focus.